%% file: main.tex
\documentclass[twoside,leqno,twocolumn]{article}

\usepackage[letterpaper]{geometry}
\usepackage{graphicx}
\usepackage{graphics}
\usepackage{ltexpprt}
\usepackage{color,soul}
\usepackage{comment}
\usepackage{cite}
\usepackage{amsmath,amssymb,amsfonts}
\usepackage{algorithmic}
\usepackage{textcomp}
\usepackage{xcolor}
\usepackage{hyperref}

\usepackage{booktabs}
\usepackage{epstopdf} 
\usepackage{multirow}
\usepackage{subfigure}
\usepackage{xspace}
\usepackage{url}

\DeclareMathOperator*{\argmin}{arg\,min}
\newcommand{\ps}{\emph{Per Sample}}
\newcommand{\m}{\emph{Mean}}
\newcommand{\e}{Test RMSE}
\newcommand{\phy}{Physical Inconsistency}
\newcommand{\pga}{\textbf{PGA}-LSTM}
\newcommand{\pgl}{{PGL}-LSTM}

\begin{document}

\title{\Large Physics-Guided Architecture (PGA) of Neural Networks  \\ for Quantifying Uncertainty in Lake Temperature Modeling}

\author{Arka Daw \footnotemark[1] \\ \and R. Quinn Thomas \footnotemark[2] \\ \and Cayelan C. Carey \footnotemark[3] \\ \and Jordan S. Read \footnotemark[4] \\ \and Alison P. Appling \footnotemark[4] \\ \and Anuj Karpatne \footnotemark[1] } 
\date{}

\maketitle

\footnotetext[1]{Department of Computer Science, Virginia Tech}
\footnotetext[2]{Department of Forest Resources and Environmental Conservation, Virginia Tech}
\footnotetext[3]{Department of Biological Sciences, Virginia Tech}
\footnotetext[4]{U.S. Geological Survey}






\begin{abstract} 
 \small\baselineskip=9pt
 

To simultaneously address the rising need of expressing uncertainties in deep learning models along with producing model outputs which are consistent with the known scientific knowledge, we propose a novel physics-guided architecture (PGA) of neural networks in the context of lake temperature modeling where the physical constraints are hard coded in the neural network architecture. This allows us to integrate such models with state of the art uncertainty estimation approaches such as Monte Carlo (MC) Dropout without sacrificing the physical consistency of our results. 
We demonstrate the effectiveness of our approach in ensuring better generalizability as well as physical consistency in MC estimates over data collected from Lake Mendota in Wisconsin and Falling Creek Reservoir in Virginia, even with limited training data. We further show that our MC estimates correctly match the distribution of ground-truth observations, thus making the PGA paradigm amenable to physically grounded uncertainty quantification.

\end{abstract}

\input{intro.tex}

\input{background.tex}

\input{method.tex}
\input{exp_setup.tex}
\input{results.tex}

\input{conclusions.tex}

\nocite{*}

\input{main.bbl}
\end{document}

%% file: intro.tex
\section{Introduction}
Given the success of deep learning methods in commercial domains such as computer vision, speech, and natural language processing, there is a growing interest in the scientific community to unlock the power of deep learning methods for advancing scientific discovery \cite{Appenzeller16,graham2008big,jonathan2011special, sejnowski2014putting}.  Out of the many reasons fueling this interest, a primary factor is the rich ecosystem of  advanced deep learning frameworks such as Conv Nets \cite{krizhevsky2012imagenet} and long short term memory (LSTM) models \cite{hochreiter1997long} that can handle complex structures in the data common in many scientific applications. Another reason is that with algorithmic innovations such as Dropout \cite{srivastava2014dropout}, we are not only moving toward robustness in deep learning but also toward better approaches for uncertainty quantification in deep learning, e.g., using the Monte Carlo (MC) Dropout method \cite{gal2016dropout}. This is especially important in scientific problems where we need to produce uncertainty bounds in addition to point estimates, e.g., in climate change applications \cite{murphy2004quantification}. 

\begin{figure}[t]
\centering
\includegraphics[width=0.85\linewidth]{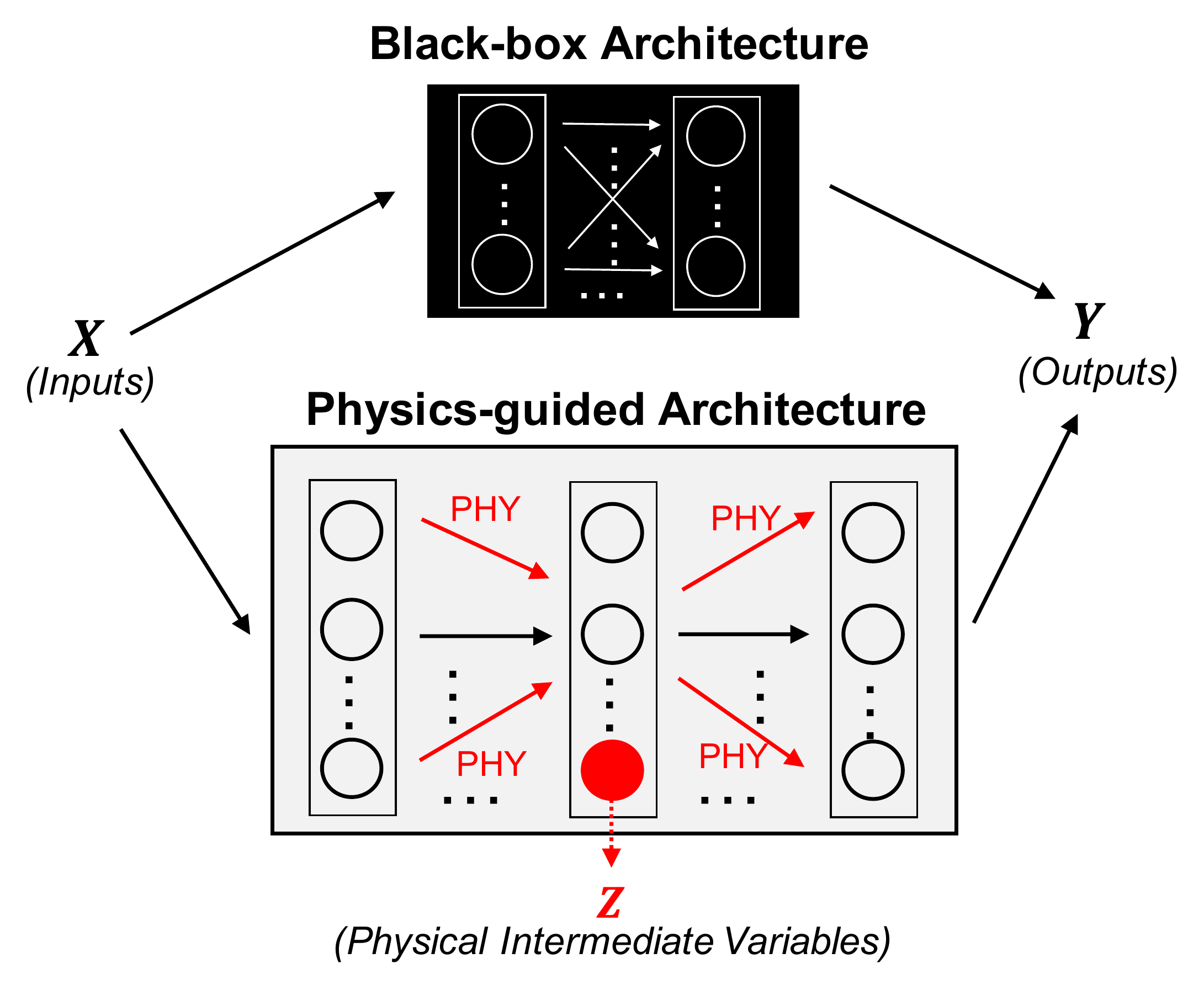}
\vspace{-2ex}
\caption{\small Physics-guided Architecture (PGA) paradigm of neural networks aims to infuse physics in neural network designs through \emph{physics-informed connections} among neurons and through \emph{physical intermediate variables}, shown in {\color{red} red}. Note: Figures in this paper are best viewed in color.}
\label{fig:pgann}
\vspace{-4ex}
\end{figure}

Despite dramatic advances in many commercial fields, current standards of deep learning has seen limited success in scientific applications (e.g., \cite{caldwell2014statistical,Lazer2014,marcus2014eight}), sometimes even leading to spectacular failures (e.g., \cite{Lazer2014}).
This is primarily because  of the \emph{black-box} nature of conventional deep learning frameworks, that are learned solely from data and are agnostic to the underlying scientific principles driving real-world phenomena. 
Since a black-box model can only be as good as the data it is fed during training, it can easily produce spurious and physically inconsistent solutions in applications suffering from paucity of labeled data.
Furthermore, dropout methods strain physical consistency further by denying some predictors and hidden states to the calculation at each iteration. Despite their value in estimating uncertainty, methods such as dropout are limited if they depart from physical realism.
As a first step in moving beyond black-box applications of deep learning, there is an emerging field of research combining scientific knowledge (or theories) with data science methods, termed \emph{theory-guided data science} \cite{karpatne2017theory}. A promising line of research in this field is to guide the learning of neural network models using \emph{physics-based loss functions} \cite{karpatne2017physics, jia2019physics, stewart2017label}, that measure the violations of physical principles in the neural network outputs. We refer to this paradigm as physics-guided learning (PGL) of neural networks. 

While PGL formulations have been shown to improve generalization performance and generate more physically consistent predictions, adding a loss function in the learning objective still does not circumvent the black-box nature of neural network architectures, involving arbitrary design choices (e.g., number of layers and nodes per layer).  As a result, black-box architectures are susceptible to producing physically inconsistent solutions with minor perturbations in the network weights, even after being trained with physics-based loss functions. This is a major concern when using uncertainty quantification methods such as MC dropout, where the network edges are randomly dropped with a small probability in the testing stage to produce a distribution of sample predictions for every test instance. Indeed, our results demonstrate that the randomness injected by MC dropout in the network weights easily breaks the ability of the PGL paradigm to preserve physical consistency in the sample predictions, leading to physically non-meaningful uncertainty estimates.

%



This paper presents innovations in an emerging field of theory-guided data science where, instead of using black-box architectures, we principally embed well-known physical principles in the neural network design. We refer to this paradigm as physics-guided architecture (PGA) of neural networks. Specifically, this paper offers two key innovations in the PGA paradigm for the illustrative problem of lake temperature modeling as illustrated in Figure \ref{fig:pgann}. First, we introduce novel \emph{physics-informed connections} among neurons in the network to capture physics-based relationships of lake temperature. Second, we associate physical meaning to some of the neurons in the network by computing physical intermediate variables $Z$ in the neural pathway from inputs to outputs. By hard-wiring physics in the model architecture, the PGA paradigm ensures physical consistency of results regardless of small perturbations in the network weights, e.g, due to MC dropout. 
We compare the efficacy of our proposed approach with baseline methods on data collected from two lakes with differing physical characteristics and climatic regimes: Lake Mendota in Wisconsin, U.S.A., and Falling Creek Reservoir in Virginia, U.S.A.

The remainder of the paper is organized as follows. Section \ref{sec:background} provides a brief background of the problem of lake temperature modeling and relevant related work. Section \ref{sec:approach} describes our proposed \textbf{PGA}-LSTM framework. Section \ref{sec:ex_setup} discusses our evaluation procedure while Section \ref{sec:results} presents results. Section \ref{sec:analysis} provides detailed analysis of our results and Section \ref{sec:conclusion} provides concluding remarks and directions for future research.



%% file: background.tex
\section{Background and Related Work}
\label{sec:background}
\subsection{Lake Temperature Modeling:}
\label{sec:lakemodeling}

\begin{figure}[tb]
\centering
\subfigure[Temperature--Density Physics]{\label{fig:temp-density} \includegraphics[width=0.26\textwidth]{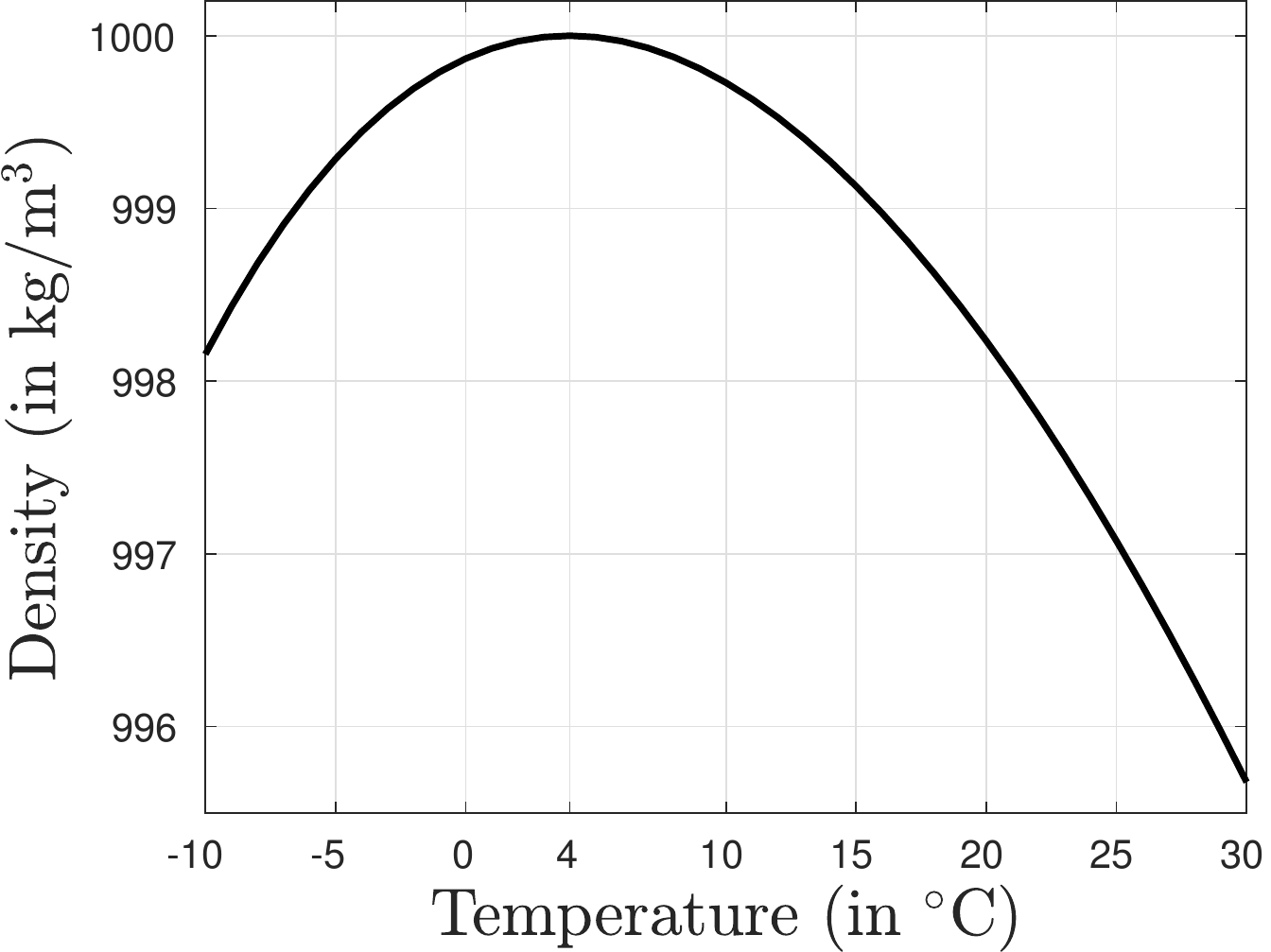}}
\subfigure[Density--Depth Physics]{\label{fig:density-depth} \includegraphics[width=0.2\textwidth]{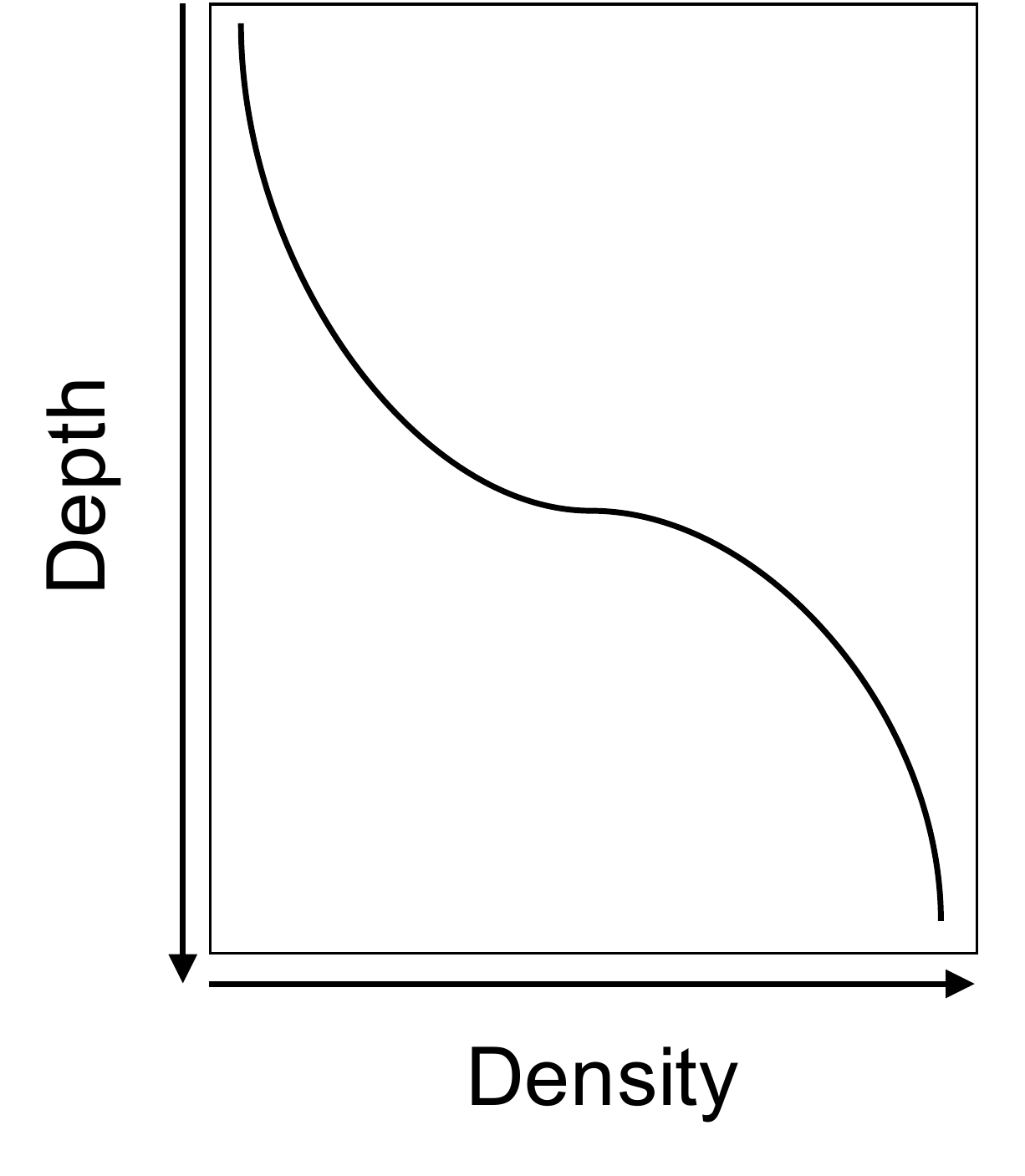}}
\vspace{-2ex}
\caption{\small Plots of two key physical relationships in lake temperature modeling: temperature-density physics and density-depth physics.}
\label{fig:phy_rel}
\vspace{-2ex}
\end{figure}

Modeling the temperature of water in a lake is important from both economic and ecological perspectives.
Water temperature is known to be principal driver of the growth, survival, and reproduction of economically viable fish \cite{roberts2013fragmentation,magnuson1979temperature} (see Appendix for more details). 
Increases in water temperature are also linked to the occurrence of aquatic invasive species \cite{rahel2008assessing,roberts2017nonnative}, which may displace fish and native aquatic organisms, and further result in harmful algal blooms \cite{harris2017predicting,paerl2008blooms}.
Hence, accurate and timely information about water temperature is necessary to monitor the ecological health of lakes and forecast future populations of fish and other aquatic taxa. 

Since observations of water temperatures are incomplete at broad spatial scales (or non-existent for most lakes), physics-based models of lake temperature, e.g., the General Lake Model (GLM) \cite{hipsey2019general}, are commonly used for studying lake processes. 
A standard formulation in these models is to assume horizontal heterogeneity is limited and that the most relevant dynamics are captured in the vertical dimension of the lake, thereby modeling the lake as a series of vertical layers. 
These modeling studies often use temperature of water at the centre of a lake at varying depth values\footnote{Depth is measured in the direction from lake surface to lake bottom.} and time points for model validation. 
We adopt the same formulation to model the temperature of water in a lake, $Y_{d,t}$ at depth $d$ and time $t$. In particular, we leverage two key physical principles of our problem to guide neural network approaches, briefly described in the following.

\vspace{1ex}

\textbf{a) Temperature-Density Physics:}
The temperature $Y$ and density $Z$ of water are non-linearly related according to the following known physical equation \cite{martin1998hydrodynamics}:
\begin{equation}
\small
Z = 1000 \times \Big( 1 - \frac{(Y + 288.9414) \times (Y - 3.9863)^2}{508929.2 \times (Y + 68.12963) } \Big)
\label{eq:temp-density}
\end{equation}
Figure \ref{fig:temp-density} shows a plot of this relationship, where we can see that water is maximally dense at 4$^\circ$C. We can use this physics to directly map temperature to density.

\vspace{1ex}
\textbf{b) Density--Depth Physics:}
The density of water $Z$ monotonically increases with depth $d$ as shown in the example plot of Figure \ref{fig:density-depth}, since denser water is heavier and goes down to the bottom of the lake. Formally,
\vspace{-2ex}
\begin{equation}
\small
    Z_{d_2} \geq Z_{d_1} \quad \text{if}~ d_2 > d_1.
    \label{eq:density-depth}
\end{equation}
These two physical relationships serve as the basis of the PGA innovations proposed in this paper for lake temperature modeling.

\subsection{Physics-guided Machine Learning:}

Physics-guided Learning (PGL) is a recent paradigm in learning neural networks \cite{karpatne2017physics, jia2019physics,stewart2017label} where along with considering the prediction loss in the target space $Y$, we also measure the violations of physical principles in the model outputs $\widehat{Y}$, represented as \emph{physics-based loss} in the PGL objective:
\vspace{-1ex}
\begin{equation}
\small
   \argmin  ~ Loss(Y,\widehat{Y}) + \lambda_{PHY} ~ \text{PHY}.Loss(\widehat{Y}).
\end{equation}
\par where $\lambda_{PHY}$ is a trade-off hyper-parameter that decides the relative importance of minimising the physical inconsistency compared to the empirical loss and the model complexity. 
By using physics-based loss, PGL restricts the search space of neural network weights to physically consistent options, thereby aiming to achieve more generalizable and physically relevant predictions.  For the problem of lake temperature modeling, Karpatne et al.\cite{karpatne2017physics} developed a PGL framework to measure violations of the two physical relationships introduced in Section \ref{sec:lakemodeling}. 
Jia et al. \cite{jia2019physics} extended this to work with time-based LSTM architectures and implemented an additional physics-based loss term to incorporate energy conservation. Other recent works like Xu et al. \cite{xu2017semantic} integrated probabilistic logic with neural networks using semantic loss for classification tasks. Also, Pathak et al. \cite{pathak2015constrained} leveraged a set of linear constraints as loss functions for weakly supervised segmentation. Marquez et al. \cite{marquez2017imposing} imposed hard constraints on neural networks using constrained optimization formulations.

A major limitation of the PGL paradigm is that the choice of the neural network architecture is still black-box and not informed by physics. Even though minimizing physics-based loss helps in physically constraining the search space of neural network weights during training, there are no architectural constraints in the neural network design that guarantee the model predictions to be physically consistent on unseen test instances. 


Physics-guided architecture of neural networks has recently gained popularity in several domains.  Leibo et al. \cite{leibo2017view} proposed network connections to incorporate Hebbian rule of learning in neuroscience for view-tolerant facial detection. Another line of work has explored ways of embedding various forms of invariance in neural networks for problems in molecular dynamics \cite{anderson2019cormorant} and turbulence modeling \cite{ling2016reynolds}.
However, none of these developments are directly applicable to our problem of temperature prediction, where we need to encode physics available in the form of monotonic relationships and presence of intermediate variables.  


\subsection{Uncertainty Quantification:}

Uncertainty quantification (UQ) is critical for model evaluation in a number of scientific applications, where rather than producing point estimates of the target variable, it is preferred to have a distribution of the possible values. In our problem of lake temperature modeling, we wish to perform UQ to ascertain the amount of confidence we can place in our temperature predictions and its estimated impact on the population of fish species and other ecological variables. 

A standard approach for performing UQ in neural networks is by using dropout \cite{srivastava2014dropout} on the trained neural network weights in the testing phase, to produce Monte Carlo samples of the target variable for every test instance---a technique called Monte Carlo (MC) dropout \cite{gal2016dropout}. 
While there are other methods in Bayesian deep learning for UQ that directly estimate posterior probabilities using priors on network weights \cite{fortunato2017bayesian},
they are generally slower than MC Dropout. 
We use MC Dropout in our approach to perform UQ for lake temperature modeling, although our proposed PGA innovations are generic and can be coupled with any other method for UQ in deep learning.

Note that every dropout network represents a slightly perturbed version of the trained ANN model. Ideally, we  want every dropout network to produce physically consistent simulations of the target variable, so that the UQ analysis is physically meaningful. However, if we use black-box architectures, we can easily obtain dropout networks that produce physically inconsistent solutions. This is because even the small amount of randomness injected by the dropout procedure may be sufficient to \emph{unlearn} the physical consistency learned during training by the PGL paradigm. In contrast, by infusing physics directly in the neural network architecture, our proposed PGA paradigm has a better chance of ensuring physical consistency in every MC dropout sample.

%% file: method.tex
\section{Proposed Framework}
\label{sec:approach}


\subsection{Overview of \textbf{PGA}-LSTM:}

\begin{figure}[t]
\vspace{-10ex}
  \includegraphics[width=\linewidth]{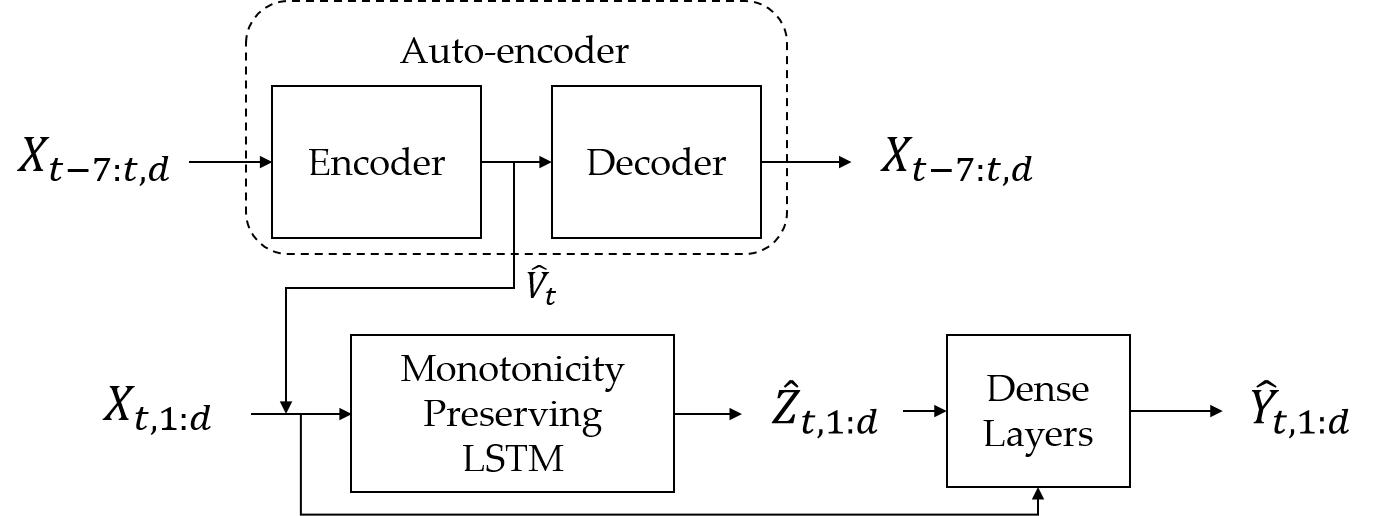}
  \caption{\small Proposed \textbf{PGA}-LSTM framework.}
  \label{fig:pgalstm}
\vspace{-4ex}
\end{figure}

Figure \ref{fig:pgalstm} provides an overview of our proposed physics-guided architecture of LSTM (\textbf{PGA}-LSTM) for lake temperature modeling. It is comprised of three basic components: (i) An LSTM based auto-encoder framework which extracts temporal features $\widehat{V_t}$ from the data at a given time $t$, (ii) a \emph{monotonicity-preserving} LSTM which uses $\widehat{V_t}$  along with additional depth-based features $X_{t,1:d}$ to predict an intermediate physical quantity: density $\widehat{Z}_{t,d}$, while ensuring that $\widehat{Z}_{t,d} \geq \widehat{Z}_{t,d-1}$, and (iii) a multi-layer perceptron model which combines the density predictions $\widehat{Z}_{t,d}$ with the input drivers $X_{t,d}$ to finally predict the temperatures $\widehat{Y}_{t,d}$. In the following, we describe these three components in detail and present an end-to-end learning procedure for the complete \textbf{PGA}-LSTM framework.

\subsection{Temporal Feature Extraction:}
The problem of lake temperature modeling can be viewed as a spatio-temporal sequential prediction problem. In order to develop a model which addresses both of these aspects simultaneously, we propose a simple yet effective method to incorporate spatio-temporal relationships into our model. 


\begin{figure}[b]
\vspace{-4ex}
\centering
  \includegraphics[width=0.9\linewidth]{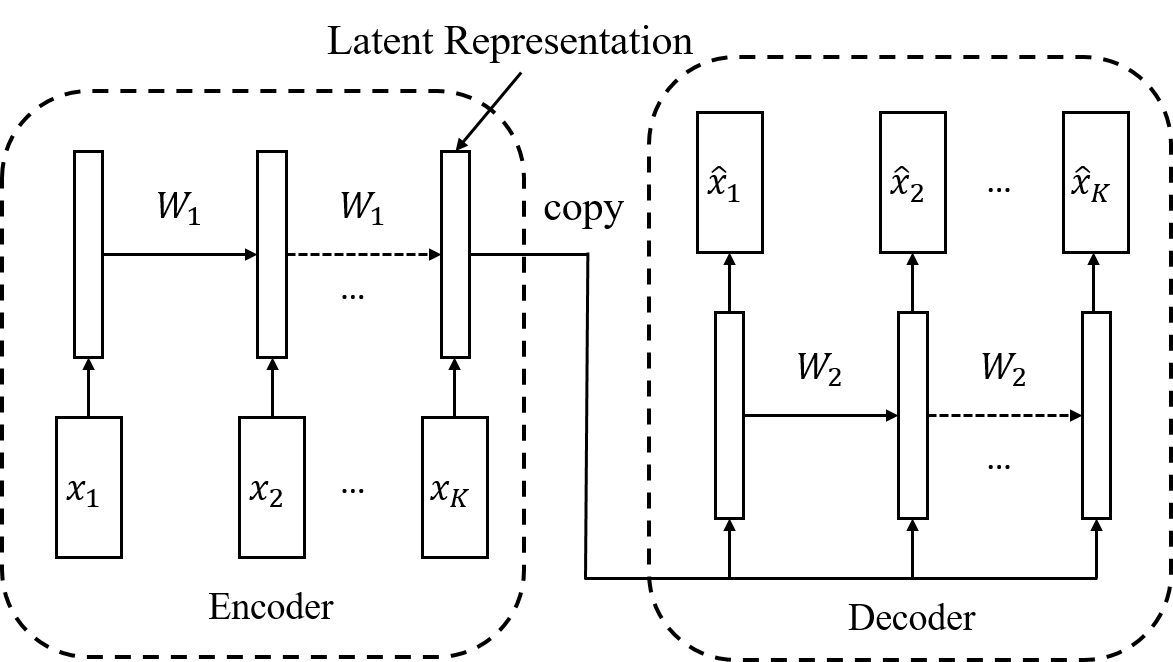}
  \vspace{-2ex}
  \caption{\small Proposed way for temporal feature extraction}
  \label{fig:autoencoder}
 \vspace{-2ex}
\end{figure}

The autoencoder consists of two Recurrent Neural Networks (RNN), the encoder LSTM and the decoder LSTM, as shown in Figure \ref{fig:autoencoder}. 
We construct an input sequence by augmenting the feature vectors of the last $7$ days of target date with its feature vector. This input $X_{t-7:t,d}$ is then fed into the encoder which generates some hidden representations $\widehat{V_t}$  for the target date. The decoder is then asked to reconstruct the entire input sequence from just the hidden representation $\widehat{V_t}$  of the target date. In order to do so, the representation must retain information about the sequential nature of the input data corresponding to the last week. 
Note that the dimensionality of the hidden representations is intentionally kept smaller than the input dimensionality. This design of auto-encoder is inspired by the earlier work of Srivastava et al. \cite{Srivastava:2015:ULV:3045118.3045209}.

\subsection{Monotonicity-preserving LSTM:}
We build upon the basic LSTM architecture \cite{hochreiter1997long} that is designed to capture long-term and short-term memory effects in predicting a target sequence $Z_{1:d}$ given an input sequence $X_{1:d}$ using a recurrent neural network (RNN) framework. The basic idea of LSTM is to remember information for arbitrary long intervals by maintaining memory cell states, ${C_d}$, and hidden states, $H_d$. The cell state $C_d$ is operated on by two neural network modules (or gates): \emph{input} gate and \emph{forget} gate, which can add or delete information in the cell state, respectively, and are learnable functions of the features, $X_d$, and the hidden state at the previous index, $H_{d-1}$. The cell state in turn affects the hidden state $H_d$ using a learnable \emph{output} gate. Finally, $H_d$ is mapped to estimates of the target variable $\widehat{Z}_d$ using a stack of fully-connected dense layers.

\begin{figure}[bt]
 \vspace{-5ex}
  \includegraphics[width=\linewidth]{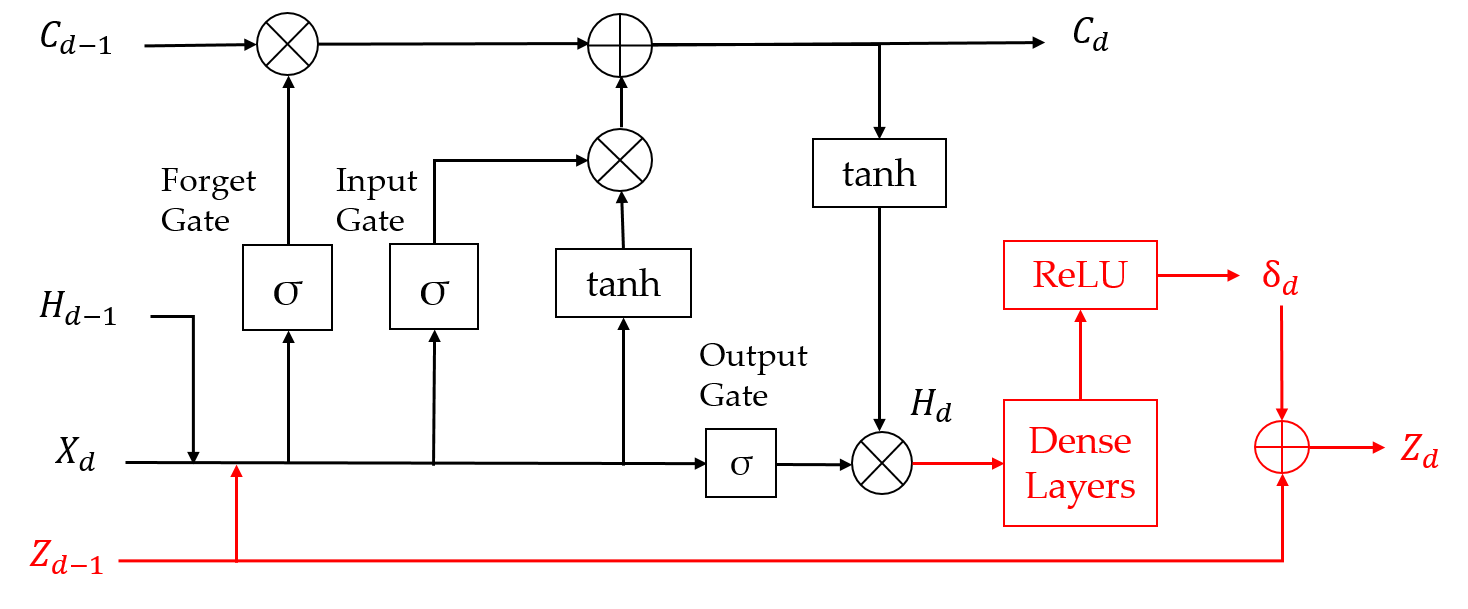}
  \caption{Monotonicity-preserving LSTM Architecture. Components in {\color{red} red} represent novel physics-informed innovations in LSTM.}
  \label{fig:mplstm}
  \vspace{-2ex}
\end{figure}

While LSTM explicitly captures recurrence relationships between hidden states at consecutive indices, $H_d$ and $H_{d-1}$, and thus offers some smoothness in its predictions, the choice of the recurrence forms are quite arbitrary and not informed by physics. In our \textbf{PGA}-LSTM formulation, $\widehat{Z}_{1:d}$ corresponds to a physically meaningful intermediate quantity: the density of water, which we know from Section \ref{sec:lakemodeling} can only increase or remain constant with depth $d$. To incorporate this knowledge of density-depth physics, we introduce novel \emph{physics-informed} connections in LSTM to have a monotonic recurrence relationship between $Z_d$ and $Z_{d-1}$. The proposed monotonicity-preserving LSTM architecture is shown in Figure \ref{fig:mplstm} and described in the following.

The main innovation in our proposed architecture (marked as red in Figure \ref{fig:mplstm}) 
is to not only keep track of the hidden state and cell state, $H_d$ and $C_{d}$, respectively, but also the physical intermediate variable, $Z_d$, which we know can only increase with depth. Hence, we consider the problem of only predicting the positive increment in density, $\delta_d$, as a function of $H_d$, which when added to $Z_{d-1}$ yields  $Z_d$. In particular, we apply a stack of $k$ dense hidden layers on $H_d$ and pass the outputs through a ReLU activation function to predict positive values of $\delta_d$. The complete set of equations for the forward pass of monotonicity-preserving LSTM is given by:



\vspace{-3ex}
\begin{equation}
\small \nonumber
\begin{aligned}
I_{d} &= \sigma(W_{i}[X_{d}, H_{d-1}, \mathbf{\color{red}Z_{d-1}}] + b_{i})  \\
F_{d} &= \sigma(W_{f}[X_{d}, H_{d-1}, \mathbf{\color{red}Z_{d-1}}] + b_{f})\\
C_{d} &= F_{d}\circ C_{d-1} + I_{d}\circ \tanh (W_{c}[X_{d}, H_{d-1}, \mathbf{\color{red}Z_{d-1}}] + b_{c}) \\
O_{d} &= \sigma(W_{o}[X_{d}, H_{d-1}, \mathbf{\color{red}Z_{d-1}}] + b_{o})\\
H_{d} &= O_{d}\circ \tanh (C_d) \label{eq:intermediate_output}\\
\mathbf{\color{red}L_{d}^{1}} &\color{red}= \mathbf{\color{red}\text{Activation} ( W_{1}  H_{d} + b_{1} )} \\
\mathbf{\color{red}L_{d}^{i}} &\color{red}= \mathbf{\color{red}\text{Activation} ( W_{i} L_{d}^{i-1} + b_{i})} \text{, where $i$ = 2,...,k.} \\
\mathbf{\color{red}\delta_{d}} &\color{red}= \color{red}\textbf{ReLU} ( W_{\delta} L_{d}^{k} + b_{\delta}) \\
\mathbf{\color{red} Z_{d}} &\color{red}= \mathbf{\color{red}Z_{d-1} + \delta_{d}} 
\end{aligned}
\end{equation}

where, terms highlighted in {\color{red} red} represent novel physics-informed innovations introduced in our proposed architecture compared to conventional LSTMs. Note that $\sigma(.)$ denotes the sigmoid activation, $\circ$ denotes the Hadamard product, $[a,b]$ denotes the concatenation of $a$ and $b$, and $(W,b)_q$ denotes learnable weight and bias terms for all values of $q$. 

While the physics-informed innovations in our \textbf{PGA}-LSTM model were specifically motivated by the density-depth physics in our target application, the idea of preserving monotonicity in LSTM outputs is useful in many other scientific applications. 
In general, our \textbf{PGA}-\emph{monotonicity-preserving} LSTM framework can be used in  applications where a target variable obeys monotonic constraints.

\subsection{Mapping Density to Temperature:}
 Having computed density as an intermediate variable in our \textbf{PGA}-LSTM framework, mapping estimates of density $\widehat{Z}_d$ at depth $d$ to estimates of temperature $\widehat{Y}_d$ at depth $d$ appears quite straightforward. Ideally, one can refer to the temperature-density physics introduced in Section \ref{sec:lakemodeling} to infer density given temperature. However, the physical mapping from density to temperature is one-to-many and thus non-unique (see Figure \ref{fig:temp-density}). In particular, a given value of density $Z$ can be mapped to two possible values of temperature $Y$, one corresponding to the freezing phase ($Y < 4^\circ C$) and the other corresponding to the warming phase ($Y > 4^\circ C$). To address this, we learn the mapping from $\widehat{Z}_d$ to $\widehat{Y}_d$ directly from the data, by concatenating $X_d$ with $\widehat{Z}_d$ and feeding the concatenated values to a stack of fully-connected dense layers with a single output node predicting the target variable $\widehat{Y}_d$. Since $\widehat{Z}_d$ is already a strong physical predictor of $Y_d$, we do not need a deep architecture to map density to temperature and thus use a small number of hidden layers. 
 

\subsection{End-to-end Learning Procedure:}
 One of the benefits of our \textbf{PGA}-LSTM framework is that along with predicting the target variable: temperature, $\widehat{Y}$, it also produces estimates of a physical intermediate variable: density, $\widehat{Z}$, as ancillary outputs. Further, during the training stage, ground-truth observations of temperature, $Y$, can be converted to ground-truth estimates of density, $Z$, using the one-to-one physical mapping from temperature to density. Hence, we perform end-to-end training of the complete \textbf{PGA}-LSTM model by minimizing the empirical loss over both $Y$ and $Z$ in the following learning objective:
 \begin{eqnarray}
 \small 
  \argmin_{(W,b)}  & \text{Loss}(Y,\widehat{Y}) + \lambda_Z ~ \text{Loss}(Z,\widehat{Z}) + \lambda_R ~ \text{R}(W) \nonumber  \\
     \text{where,}& Loss(Y,\widehat{Y}) = \frac{1}{N} \sum_{d}\sum_{t} (Y_{d,t} - \widehat{Y}_{d,t})^2, \nonumber\\
         & Loss(Z,\widehat{Z}) = \frac{1}{N} \sum_{d}\sum_{t} (Z_{d,t} - \widehat{Z}_{d,t})^2, \nonumber \\
    &\text{R}(W) =  ||W||_2. \label{eq:loss} 
\end{eqnarray}

Here, $Y_{d,t}$ and $Z_{d,t}$ are the observed temperature and density values, respectively, at depth $d$ and time $t$, $N$ is the total number of observations, $(W,b)$ is the combined set of weights and bias terms, respectively, across all components of \textbf{PGA}-LSTM, and $\lambda_Z$ and $\lambda_R$ are the trade-off parameters for the density prediction loss and regularization loss, respectively.

%% file: exp_setup.tex
 \section{Evaluation Setup}
 \label{sec:ex_setup}

 \subsection{Data and Experiment Design:}
 Our proposed \textbf{PGA}-LSTM model was trained and tested on two lakes that differed in depth, size, and climatic conditions. The first lake, Lake Mendota in Wisconsin, USA, is approximately 40 km$^2$ in surface area with a maximum depth of approximately 25 m. Lake Mendota is a dimictic lake with seasonal variation in water temperatures from 0° in the winter to nearly 30° in the summer. Lake Mendota thermally stratifies each spring and mixes in the fall before ice cover appears, typically from late December or early January until March or April. Thermal stratification is the process by which lake surface warming generates differences in temperatures between waters closer to the surface with the colder waters below \cite{wetzel2001lake}. Temperature data for Lake Mendota were collected from an instrumented buoy situated near the deepest part of the lake. The buoy collected temperature observations every 0.5 m from surface to 2 m and every 1 m from 2 m to 20m (for a total of 23 depth locations). 
 The overall data for lake Mendota consisted of 35,213 observations. This buoy was removed from the lake in the fall and replaced in the spring to avoid ice damage. To supplement the dataset with additional temperature measurements during periods when the buoy was not operational, we added manual temperature observations from the NTL-LTER sampling program to generate a dataset from April 2009 to December 2017. As such, the temperature measurements varied both across depth and over time. 
 
  Falling Creek Reservoir (FCR) in Virginia, USA, is approximately 0.119 km$^2$ in surface area with a maximum depth of 9.3 m \cite{carey2018fcr}. Similar to Mendota, FCR is also dimictic and has seasonal variation in water temperatures from 0° in the winter to nearly 30° in the summer. FCR thermally stratifies each spring and mixes in the fall. Ice cover will occur at FCR each winter, but the duration of ice varies substantially year to year, from a few weeks to multiple months, depending on winter weather. Water temperature data for FCR were collected between 2013 and 2018 using manual casts of a CTD (Conductivity, Temperature, and Depth) SeaBird profiler at the deepest site of the reservoir. CTD profiles were generally collected weekly to subweekly during April to October, monthly in October - December, and then intermittently in the winter months.  
  The CTD profiler collects observations every 4 Hz as it is lowered through the water column, resulting in ~1 cm resolution data. 
To standardize data among CTD profiles over time, the water temperature data were discretized to every 0.33 meter for a total of 28 measurement depths. The overall data for FCR consisted of 7588 observations \cite{carey2019time}.

For both lakes features consisted of: day of year, depth, air temperature, shortwave radiation, longwave radiation, relative humidity, wind speed, rain, growing degree days, if the lake was frozen, and if it was snowing. With the exception of depth, all driver data were measured or calculated from meteorological datasets, and thus remained constant for a particular time $t$ across all depths. 
Additional simulated water temperature output from a physics-based model (GLM), using the features above as inputs, were also used as features.  
 
 We partitioned the data into two contiguous time windows to be used for training and testing, such that there is no temporal auto-correlation between the training and test sets. 
 The first 4 years was used for training for both the lakes and the remaining for testing. For the training subset selection, we have randomly selected dates and cumulatively added the number of observations for each date till we reached the required number of observations given a particular training fraction. 
 Both the input and density outputs were normalised to zero mean unity standard deviation. The main temperature output $Y$ was not normalised
\footnote{The codes for \textbf{PGA}-LSTM are available on Github: \href{https://github.com/arkadaw9/PGA_LSTM}{https://github.com/arkadaw9/PGA\_LSTM}}.
 
 \subsection{Model Specifications}
 
 We have used previous 7 days of data to extract a 5 dimensional temporal embedding for each date. 
It is well known that a LSTM keeps on updating its memory state from the previous cell state to the next, i.e. it learns the history and uses that information to predict more accurately. Thus, LSTM needs a few memory cells to build up its memory in order to predict outputs for the full sequence. 
 For our proposed monotonicity-preserving LSTM we perform padding to handle this shortcoming of LSTM, where we copy the features at the surface of the lake as padding values.  A padding of size 10 was used for both lakes, and the Lake mendota and FCR was discretized into 50 and 28 depth intervals respectively. The number of units for the \emph{monotonicity-preserving} LSTM recurrent units was set to 8, followed by two dense layers with 5 hidden neurons and  `ELU' activations each with a final single neuron layer to predict the $\delta_{d}$ value. The $\delta_{d}$ values are then used to update intermediate physical variable $Z_{d}$. The second component of the \textbf{PGA}-LSTM, which maps $Z$ into $Y$, comprises of another set of two dense layers, again each with 5 hidden neurons and 'ELU' activations. A dense layer with only one neuron was used to predict the final temperature.
 To obtain uncertainty estimates, the dropout method was used during the testing phase \cite{gal2016dropout}. 
 We have used a dropout probability of 0.2, and for each input we have randomly created 100 different dropout networks to get a distribution on the model outputs.

 \subsection{Evaluation Metrics:}
 We consider the root mean square error (RMSE) of a model on the test set as a metric of generalizability of the model. We also consider \phy{} as another evaluation metric, which is defined as the fraction of times the MC sample predictions at consecutive depths are physically inconsistent, i.e., they violate the density-depth relationship. We have used a tolerance value of $10^{-5}$ kg/m$^3$ to decide if a difference in density across consecutive depths is physically inconsistent or not. 
 \subsection{Baselines:}\label{baselines}
 We have chosen the following baselines for comparison. (1) An LSTM with similar architecture as a \textbf{PGA}-LSTM was chosen for comparison. The black-box LSTM has 8 memory units followed by four dense layers with 5 hidden units each, finally followed by a dense layer with one unit.  (2) The PGL-LSTM was used as another baseline with a similar architecture to the LSTM but using physics-guided learning in the form of loss functions. The PGL-LSTM tries to minimise the physics based loss that can evaluated by computing the physical inconsistency of two consecutive depth outputs. Note that we did not consider the PGRNN approach \cite{xiaowei_sdm} for lake temperature modeling using physics-guided learning in RNNs as a baseline in this paper due to two main reasons. First, this paper only considers density-depth relationships while PGRNN also considers energy conservation. Second, PGRNN builds RNNs in time dimension whereas we are building LSTM in the depth dimension, to make use of the density-depth relations directly in the creation of physics-guided architecture.
 
 

 

%% file: results.tex
\section{Results}
\label{sec:results}

\subsection{Comparing PGA-LSTM with Baselines:}

\begin{table}[t]
\centering
\resizebox{\columnwidth}{!}{
\begin{tabular}{|l|c|c|c|c|}
\hline
         & \multicolumn{2}{c|}{Test RMSE (in $^{\circ}C$)}      & \multicolumn{2}{c|}{Physical Inconsistency} \\[1ex]  \hline
         & \emph{Per Sample}        & \emph{Mean}          & \emph{Per Sample}           & \emph{Mean}                    \\[1ex] \hline
LSTM     & $2.25\pm0.14$ & $1.63\pm0.08$ & $0.32\pm0.03$     & $0.10\pm0.02$    \\[1ex]      \hline
{PGL} & $2.30\pm0.12$ & $\mathbf{1.62\pm0.10}$ & $0.34\pm0.02$    & $0.12\pm0.02$           \\[1ex] \hline
\textbf{PGA} & $\mathbf{2.09\pm0.18}$ & $1.63\pm0.04$ & $\mathbf{0.01\pm0.01}$    & $\mathbf{0.00\pm0.00}$   \\[1ex] \hline
\end{tabular}
}
\caption{ \small\baselineskip=9pt Results on Lake Mendota using 40\% training data.}

\label{tab:Mendota}
\end{table}

\begin{table}[t]
\centering
\resizebox{\columnwidth}{!}{
\begin{tabular}{|l|c|c|c|c|}
\hline
         & \multicolumn{2}{c|}{Test RMSE (in $^{\circ} C$)}      & \multicolumn{2}{c|}{Physical Inconsistency} \\[1ex]  \hline
         & \emph{Per Sample}        & \emph{Mean}          & \emph{Per Sample}           & \emph{Mean}                    \\[1ex]  \hline
LSTM     & $2.96\pm0.22$ & $2.27\pm0.17$ & $0.28\pm0.02$    & $0.07\pm0.03$    \\[1ex]      \hline
{PGL} & $2.84\pm0.16$ & $2.12\pm0.13$ & $0.27\pm0.02$     & $0.08\pm0.03$           \\[1ex] \hline
\textbf{PGA} & $\mathbf{2.19\pm0.21}$ & $\mathbf{1.88\pm0.12}$ & $\mathbf{0.00\pm0.01}$    & $\mathbf{0.00\pm0.00}$   \\[1ex] \hline
\end{tabular}
}
\caption{ \small\baselineskip=9pt Results on Falling Creek Reservoir using 40\% training data.}
\label{tab:fcr}
\vspace{-1ex}
\end{table}

Tables \ref{tab:Mendota} and \ref{tab:fcr} compare the performance of \textbf{PGA}-LSTM with baseline methods on Lake Mendota and FCR, respectively, using $40\%$ data for training on both lakes. We are interested in two evaluation metrics: Test RMSE and Physical Inconsistency of the predicted temperature profiles. Both metrics can be evaluated either on the individual MC samples (referred to as \emph{Per Sample}) or on the mean of the MC samples (referred to as \emph{Mean}). (Note that we are interested in \ps{} evaluation as we want every MC sample to be accurate and physically consistent for the \m{} results to make sense in scientific applications.) We can see from Table \ref{tab:Mendota} that on Lake Mendota, LSTM has fairly high \emph{Per Sample} Test RMSE, illustrating the limitations of black-box models in achieving good generalizability. Further, the \emph{Per Sample} Physical inconsistency of LSTM is 0.32, which indicates that the MC samples generated from LSTM are physically inconsistent 32\% of the time. Such samples, even if they match with the test labels, represent physically non-meaningful estimates that cannot be used for subsequent scientific studies including uncertainty quantification. 
If we consider the mean of MC samples generated from LSTM, we can get lower \m{} \e{}, due to the cancellation of noise through aggregation. However, the \m{} \phy{} of LSTM is still fairly high.

If we employ the PGL paradigm, we can see that {PGL}-LSTM shows little to no improvement in RMSE or Physical Inconsistency in comparison to LSTM. A more serious concern with  PGL-LSTM is that they provide little to no improvement in the physical consistency of the prediction samples, obtained via MC dropout. Note that every dropout network represents a slightly perturbed version of the trained neural network model. Ideally, we  want every dropout network to produce physically consistent simulations of the target variable, so that the UQ analysis is physically meaningful. However, if we use black-box architectures, it is highly likely to obtain dropout networks that produce physically inconsistent solutions even after using the PGL paradigm. This is because the dropout procedure effectively injects a small amount of randomness in the neural network weights, which may be sufficient to \emph{unlearn} the physical consistency introduced during training by the PGL paradigm. 
In contrast to the baseline methods, we can observe that our proposed \textbf{PGA}-LSTM model shows the smallest \ps{} \e{} while always preserving physical consistency, even after performing MC dropout. 
Note that the number of learnable parameters in \textbf{PGA}-LSTM is very similar to all baseline models. For example, LSTM and \textbf{PGA}-LSTM use the same number of features in the hidden states, $H$, and have the same number of dense hidden layers. However, the difference in \textbf{PGA}-LSTM architecture is that one of the hidden neurons in the neural network is explicitly trained to express a physically meaningful quantity: density, which is known to be a key intermediate variable in the mapping from input drivers to temperature. Further, \textbf{PGA}-LSTM explicitly encodes physics-informed connections in LSTM to maintain the monontonic recurrence between density and depth. These physics-based architectural changes ensure that the learned neural network is physically consistent and is robust to minor perturbations in the network weights, thus demonstrating better generalization power.
Table \ref{tab:fcr} shows similar trends in the results of \textbf{PGA}-LSTM w.r.t. baseline methods on FCR. 
We can see that \pga{} is able to reduce the \ps{} \e{} from 2.96$^\circ$ for LSTM to 2.19$^\circ$, which represents a significant improvement in the accuracy of MC samples generated by our proposed method. Further, since the \phy{} of \pga{} is close to 0, our proposed method produces meaningful samples that can be employed by lake scientists for subsequent analyses of lake processes that are affected by lake temperature, such as the growth and survival of fish species. By taking the mean of our MC samples, the \m{} \e{} of \pga{} further reduces to 1.88$^\circ$.  
\begin{figure}[tb]
\centering
\subfigure[\small Lake Mendota]{\label{fig:mendotarmse} 
\includegraphics[width=0.9\columnwidth]{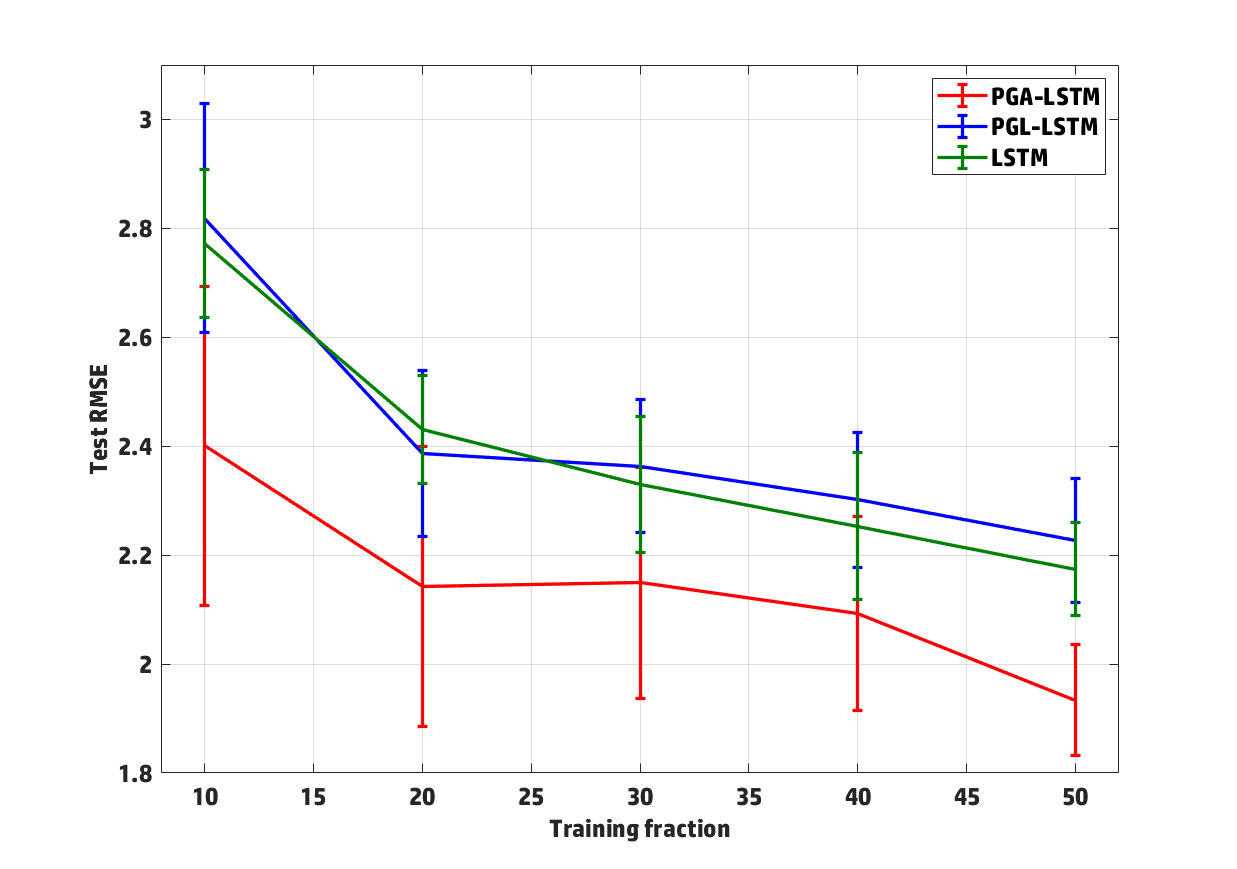}}
\\
\subfigure[\small FCR]{\label{fig:roanokermse} 
\includegraphics[width=0.9\columnwidth]{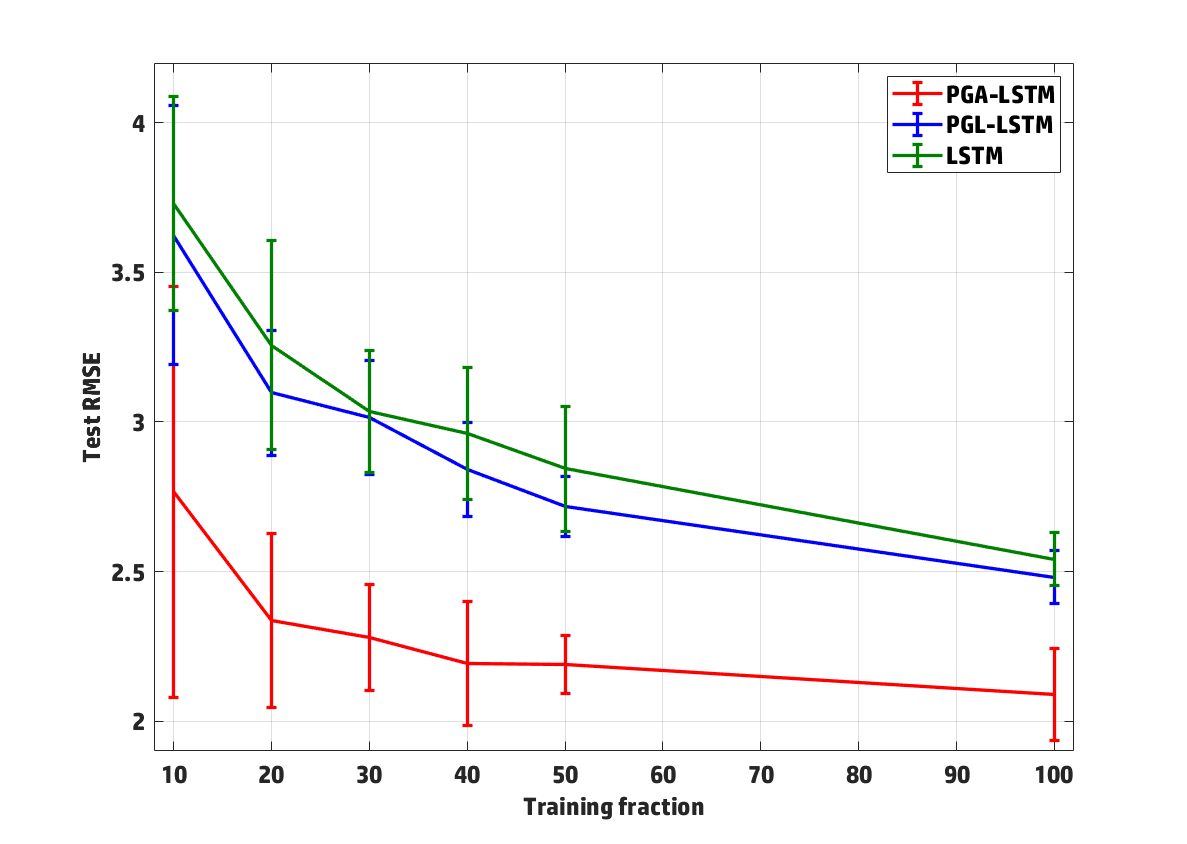}}
\caption{\small Test RMSE (\emph{per sample}) on varying training sizes.}
\label{fig:rmse_vs_tr_frac}
\vspace{-5ex}
\end{figure}

\begin{figure*}[tb]
\centering
\subfigure[LSTM Profiles (15 samples)]{\label{fig:allLSTM} \includegraphics[width=0.32\textwidth]{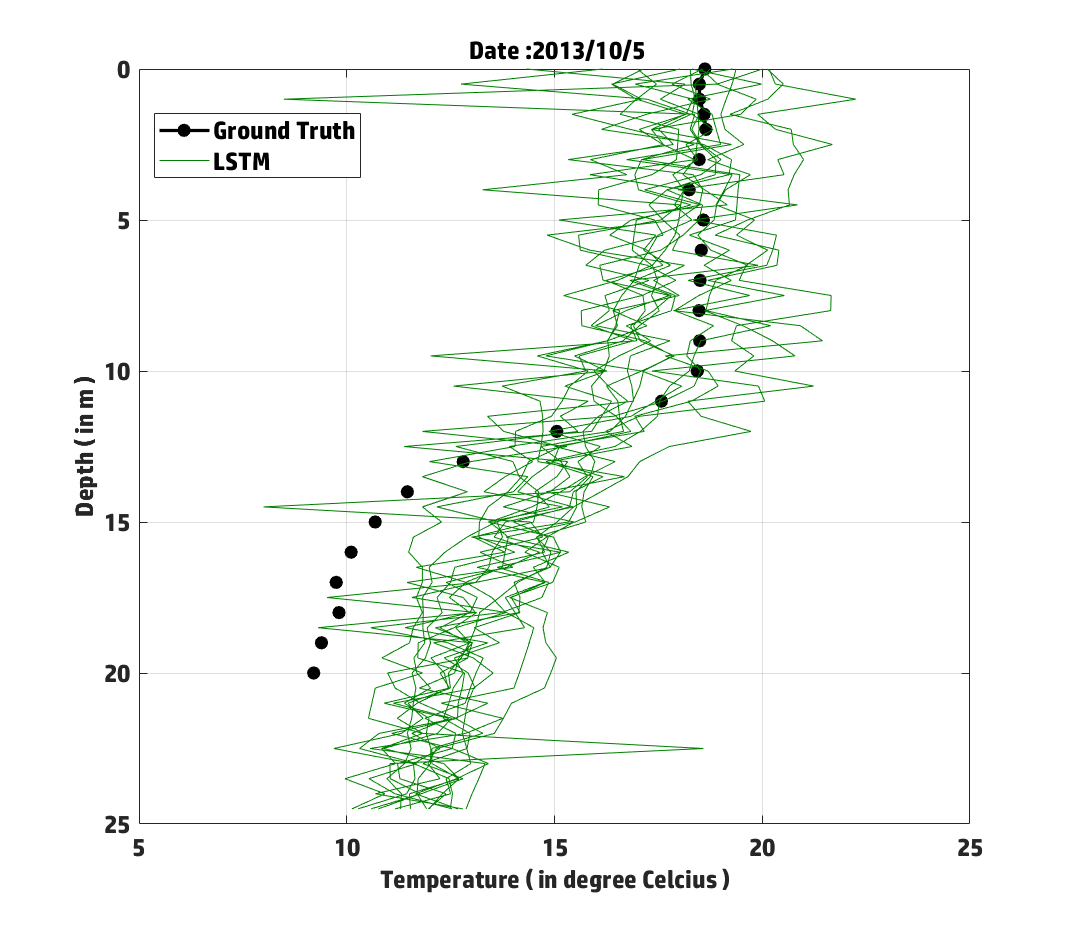}}
\subfigure[PGL-LSTM Profiles (15 samples)]{\label{fig:allPGLLSTM} \includegraphics[width=0.32\textwidth]{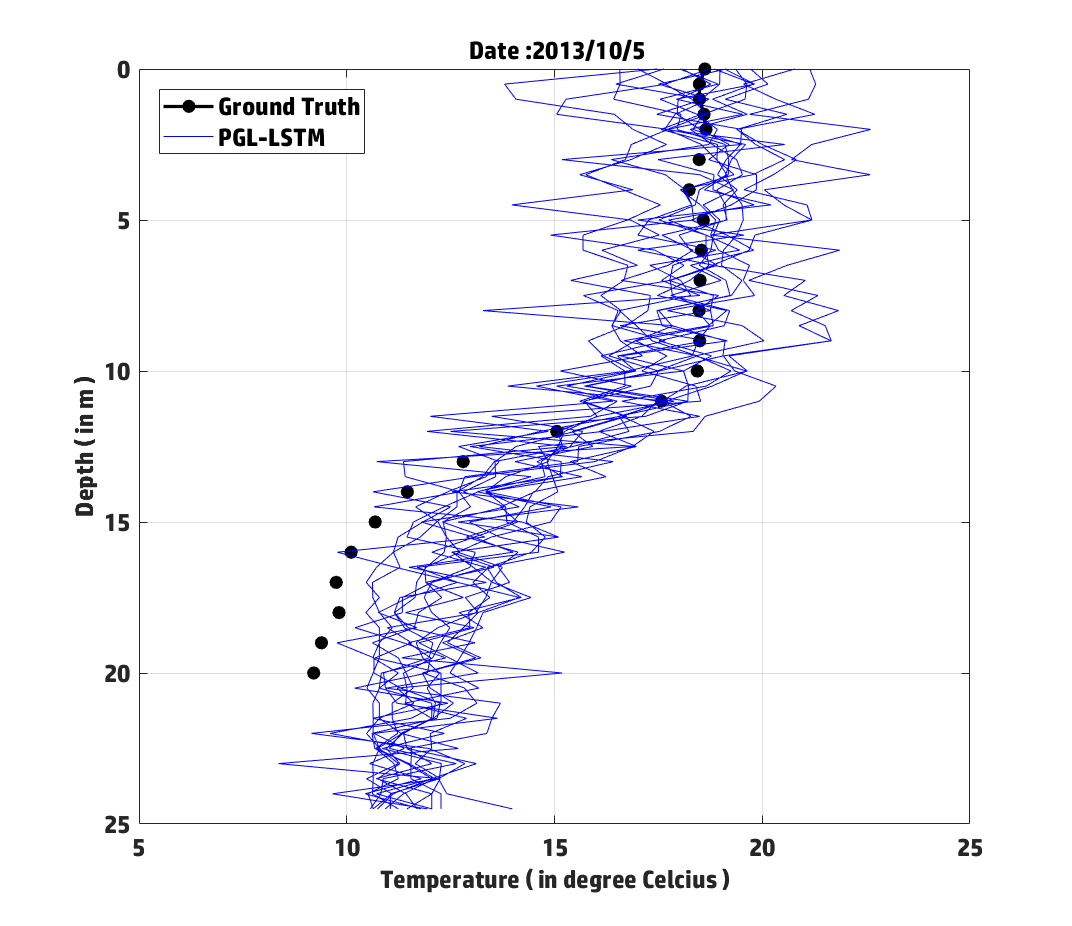}}
\subfigure[\textbf{PGA}-LSTM Profiles (15 samples)]{\label{fig:allPGALSTM} \includegraphics[width=0.32\textwidth]{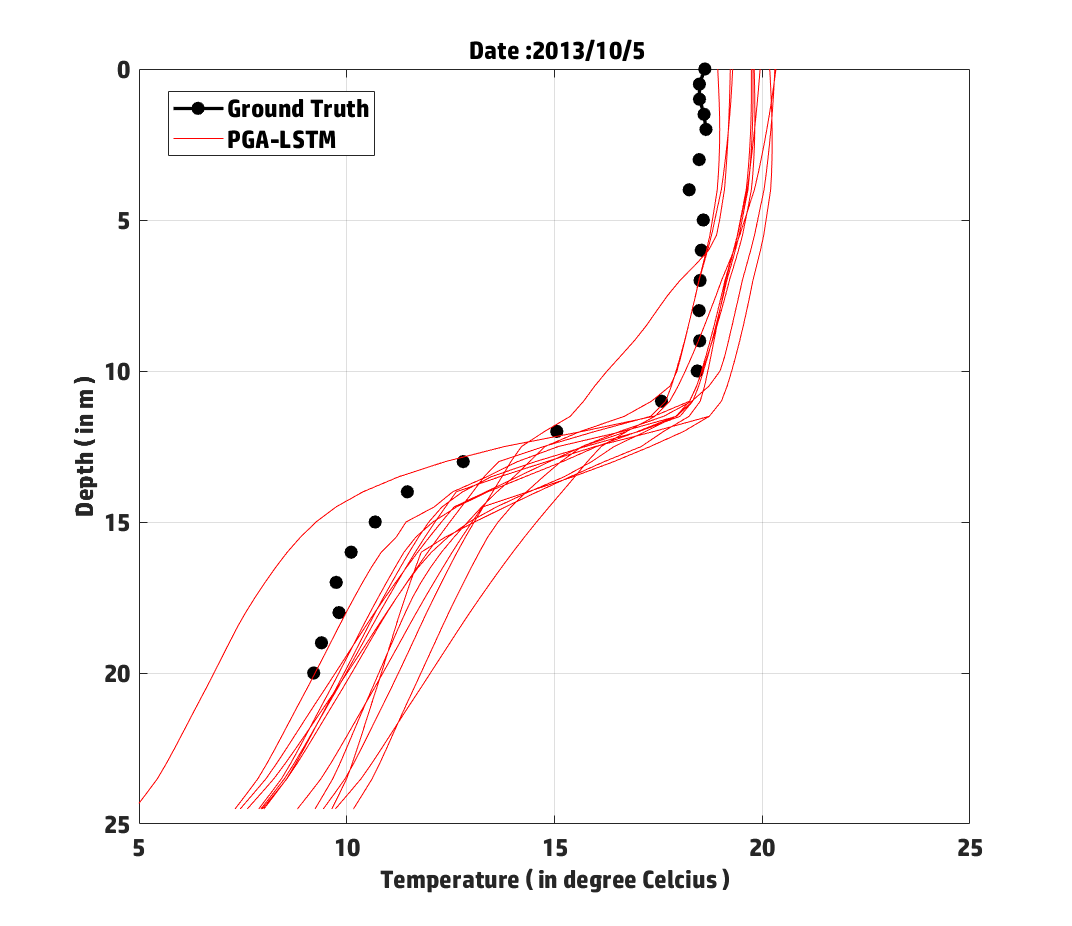}}
\\
\vspace{-1ex}
\subfigure[LSTM Mean and Variance]{\label{fig:meanLSTM} \includegraphics[width=0.32\textwidth]{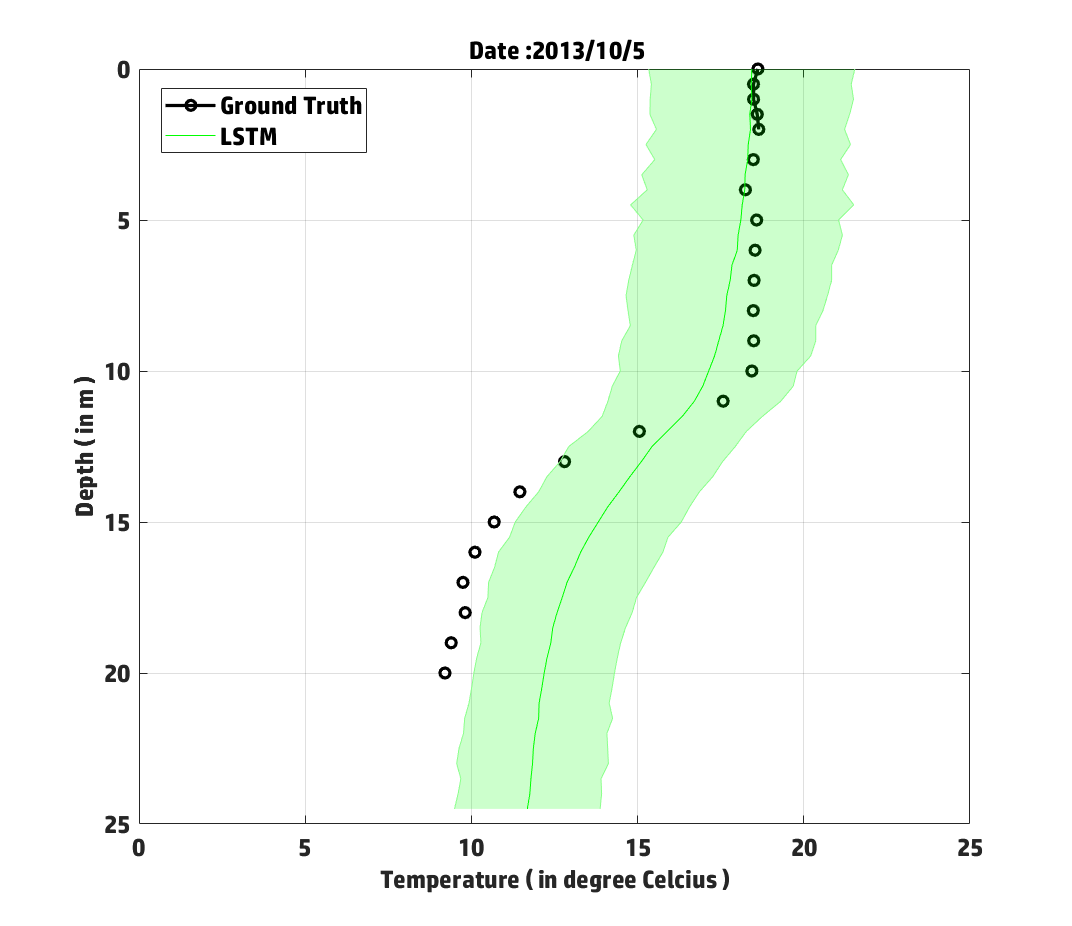}}
\subfigure[PGL-LSTM Mean and Variance]{\label{fig:meanPGLLSTM} \includegraphics[width=0.32\textwidth]{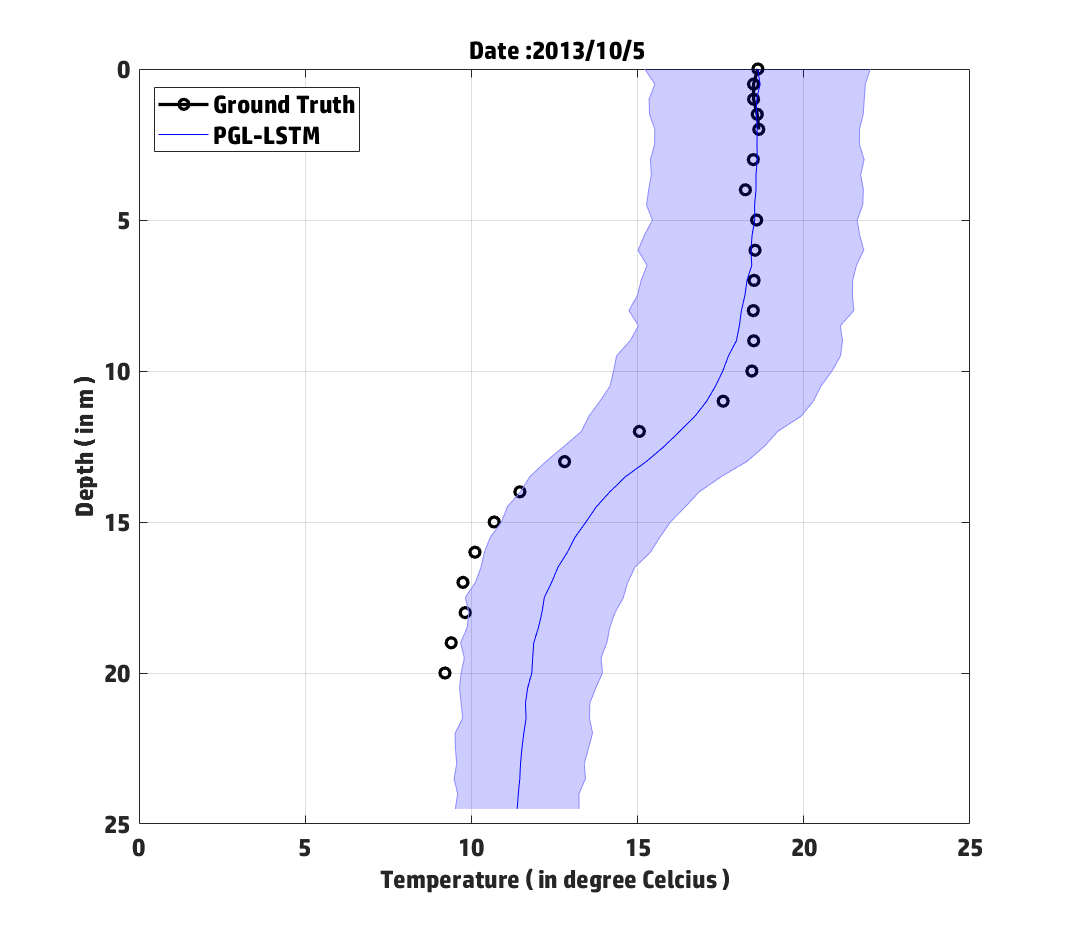}}
\subfigure[\textbf{PGA}-LSTM Mean and Variance]{\label{fig:meanPGALSTM} \includegraphics[width=0.32\textwidth]{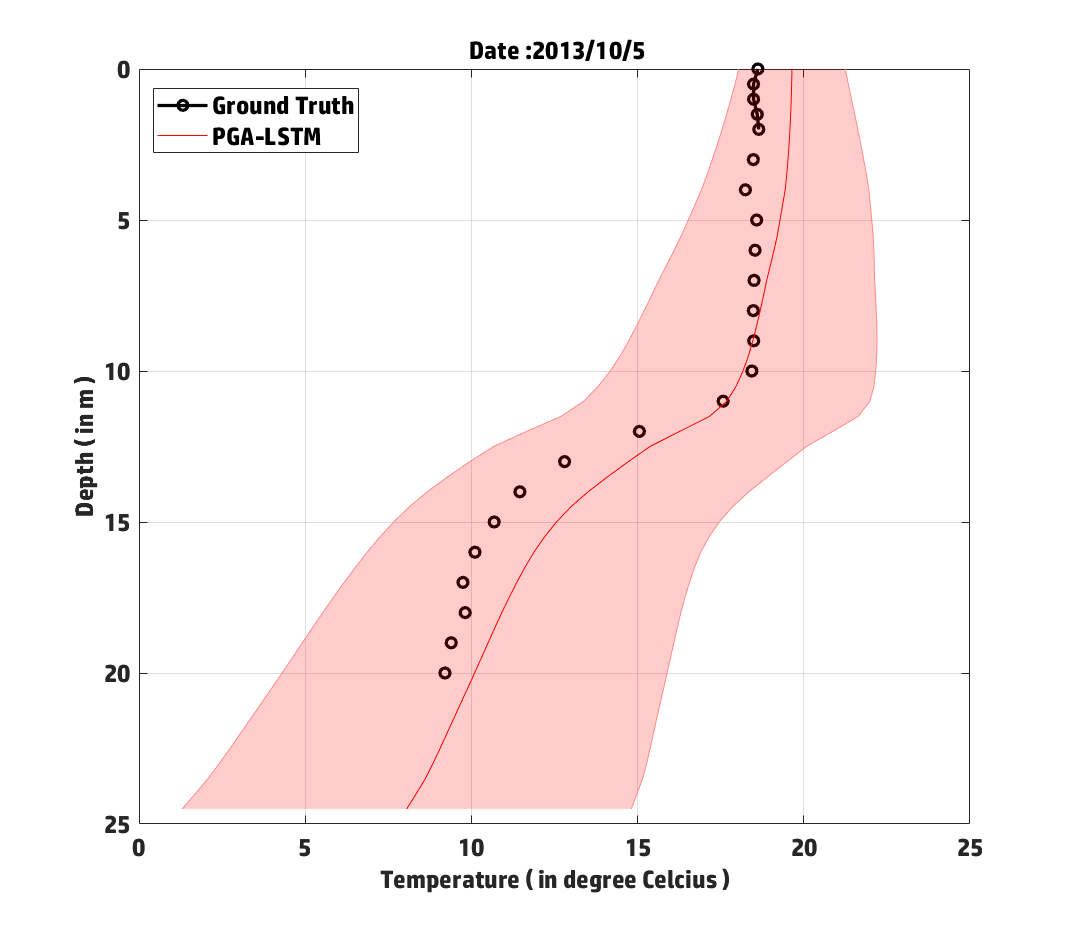}}
\vspace{-1ex}
\caption{Temperature profiles of comparative models on October 5, 2013 for Lake Mendota.}
\label{fig:profilesmendota}
\end{figure*}

\subsection{Effect of Varying Training Size:}
To demonstrate the effects of reducing training size on the accuracy of comparative models, Figure \ref{fig:rmse_vs_tr_frac} shows the \ps{} \e{} of \pga{}, \pgl{}, and LSTM on varying training fractions. We can see that the test RMSE of all methods increase as we reduce the amount of data available for training in both Lake Mendota and FCR. 
However, we can see that \pga{} shows the lowest \e{} for all values of training fractions in both the lakes.
While Lake Mendota and FCR represent heavily studied water bodies, a majority of lakes in the USA (and the world) suffer from limited number of observations. Hence, by comparing models in scarcity of training data on these lakes, we intend to simulate real-life scenarios on other unseen lakes where temperature models have to be deployed.
Note that in FCR, the rate of increase in RMSE as we reduce training size is lowest for \pga{} as compared to all baseline methods. This resonates with the fact that the need for introducing physics to achieve better test RMSE is higher at smaller training sizes, when black-box models have higher risks of over-fitting and learning spurious solutions.


\section{Analysis of Results}
\label{sec:analysis}

\subsection{Visualizing Temperature Profiles:}
Beyond simply analyzing the performance of \pga{} in terms of two evaluation metrics, here we perform further visualizations of the sample profiles of temperature predicted by our proposed method in comparison with baselines, to assess the physical validity of our results.
Figures \ref{fig:allLSTM}, \ref{fig:allPGLLSTM}, and \ref{fig:allPGALSTM} show plots of 15 sample temperature profiles generated by comparative models on a representative test date of October 15, 2013 in Lake Mendota, when trained on 40\% data. Note that these 15 samples have been selected at random from the pool of dropout MC samples generated on this test date across all 10 random runs of training. We can observe that the sample profiles of LSTM and \pgl{} are highly physically inconsistent (i.e., temperature profiles show no monotonic behavior with depth), even if they appear close to the ground-truth observations on this date. 
Hence, despite their RMSE values, a lake scientist will have lower confidence in trusting their results and using them in subsequent scientific analyses.
In contrast to baseline methods, \pga{} produces sample profiles that are always physically consistent, and thus are useful from a domain perspective.

To analyze the validity of MC profiles in capturing the uncertainty around temperature predictions, Figures \ref{fig:meanLSTM}, \ref{fig:meanPGLLSTM}, and \ref{fig:meanPGALSTM} show the mean and variance of comparative models for the complete pool of dropout samples generated on this test date across all 10 random runs of training. The error bars in these plots have been generated using two standard deviations around the mean, thus capturing $99.7\%$ of samples under Gaussian assumption. We can see that the mean profiles of LSTM and \pgl{} are close to the ground-truth and the error bars engulf the ground-truth in the shallower portion of the lake (depth $< 10$m). However, as we move to the deeper portion of the lake (depth $>10$m), the ground-truth observations start to depart from the distribution of samples generated by LSTM and \pgl{} and escapes outside the error bars. On the other hand, the distribution of samples generated using \pga{} accurately envelops the ground-truth observations at every portion of the lake irrespective of depth. \footnote{While Figure \ref{fig:profilesmendota} provides results over a single test date in Lake Mendota, videos of the results for both lakes for all test dates are available at the following link : \href{https://drive.google.com/drive/folders/1IoPYhEhsO6134HvryUDoD-CCRE7hk3dy?usp=sharing}{\textbf{click here}} }

Notice that in contrast to baseline methods, the variance of \pga{} samples gradually increases with depth. The challenge in predicting the temperatures in the deeper depths is that the set of input features of the model are not adequate. For example, improved water clarity in some years make the bottom of the lake have faster warming rates. Since we don't have water clarity as an input feature, it makes it harder for the model to predict the dynamics of the lake in deeper depths on unseen test years. Hence, by reporting higher uncertainty at deeper depths, \pga{} is exhibiting physically meaningful behavior that can be explained using domain understanding.

\subsection{Assessing Uncertainty Estimates:}
\begin{figure}
\includegraphics[width=\linewidth]{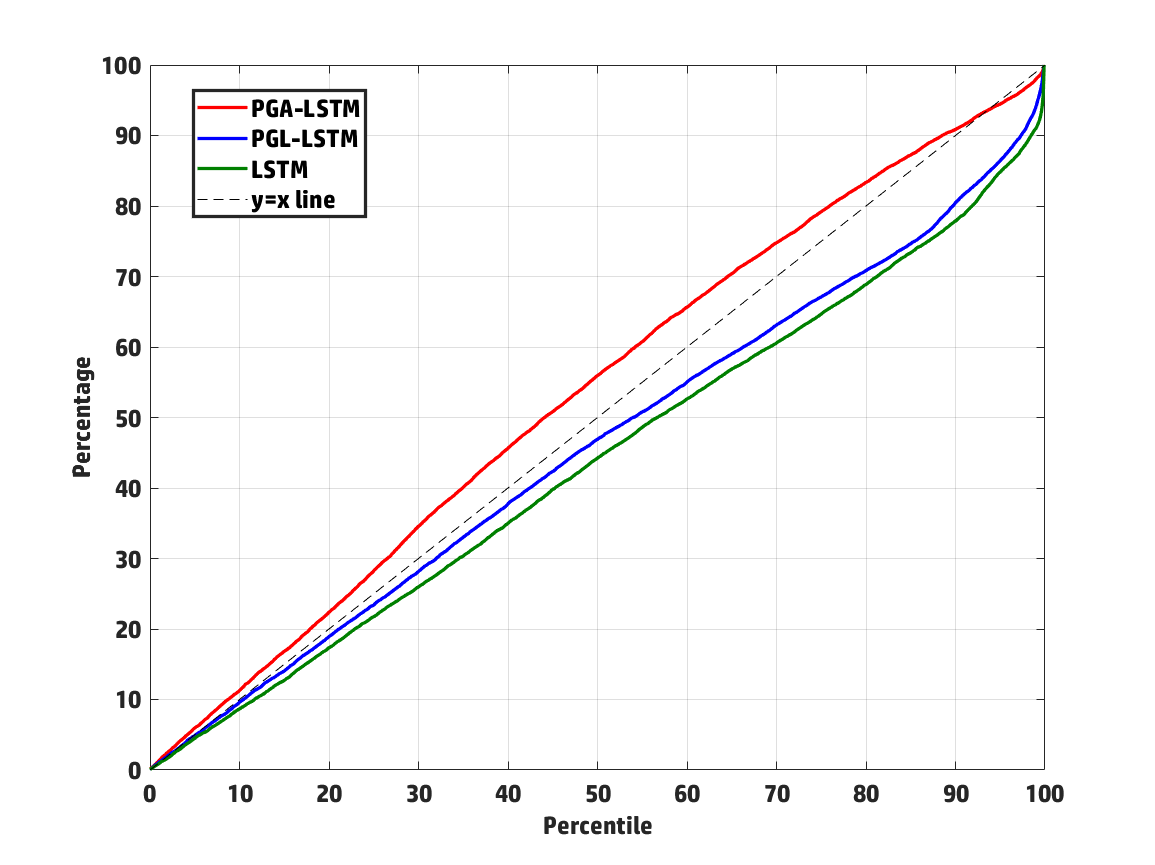}
\caption{\small Cumulative percentage of observations within a certain percentile of samples of comparative models on Lake Mendota.}
\label{fig:mendota_pp_plot}
\vspace{-4ex}
\end{figure}

While Figure \ref{fig:meanPGALSTM} shows that the ground-truth observations are contained within $99.7\%$ samples of \pga{} on a given date in Lake Mendota, we attempt to quantitatively assess the validity of our uncertainty estimates as compared to baseline methods across all dates. Ideally, if the uncertainty estimates produced by a model are valid, we should expect the distribution of its samples to accurately match the distribution of ground-truth observations on test points. In other words, if we look at the $k^\text{th}$ percentile of samples generated by an ideal model, then we should expect to observe $k\%$ of ground-truth test points to fall within it. To capture this idea, we first fit a Gaussian distribution on the complete pool of samples generated by a model on a test point, and then estimate the two-tailed percentile of the ground-truth observed at that point. Figure \ref{fig:mendota_pp_plot} plots the cumulative percentage of ground-truth observations ($Y$-axis) that fall within a certain percentile of samples generated by comparative models ($X$-axis). The ideal model is represented by the diagonal line $y =x$, where the percentage of ground-truth points within a percentile is equal to the percentile value. Models that are over-confident would have fewer ground-truth points within a certain percentile and hence would lie below the diagonal. Conversely, models that are under-confident would reside above the $y =x$ diagonal. We can see from Figure \ref{fig:mendota_pp_plot} that the baseline models, LSTM and \pgl{}, lie just below the diagonal and hence produce slightly over-confident uncertainty estimates, i.e., the distribution of ground-truth points sometimes falls outside the distribution of MC samples. On the other hand, \pga{} lies just above the diagonal and hence produces slightly larger uncertainty estimates than what is ideally expected. Note that in the absence of any information about the ground-truth on test dates, it is generally desirable to be slightly under-confident and produce wider uncertainty bounds than to be over-confident with narrower bounds. Further, even though the uncertainty bounds of \pga{} are wider, it produces lower \ps{} \e{} than all other baseline methods. This illustrates that \pga{} produces reasonably good uncertainty estimates that captures the distribution of the ground-truth observations.

%% file: conclusions.tex
\section{Conclusions and Future Work}
\label{sec:conclusion}

This paper explored an emerging direction in theory-guided data science to move beyond black-box neural network architectures and design physics-guided architectures (PGA) of neural networks that are informed by physics. We specifically develop a novel \textbf{PGA}-LSTM model for the problem of lake temperature modeling, where we design a \emph{monotonicity-preserving} LSTM module to predict physically consistent  densities. We compared our \textbf{PGA}-LSTM model with baseline methods to demonstrate its ability to produce generalizable and physically consistent solutions, even after making minor perturbations in the network weights by the Monte Carlo (MC) dropout method for uncertainty quantification.

Future work will explore applications of the proposed \textbf{PGA}-LSTM in other scientific problems that show monotonic recurrence relationships. Future work will also explore the effect of other state of the art uncertainty prediction methods on physics-guided architecture models. Extensions of the PGA framework for capturing more complex forms of physical relationships in space and time will be explored as well. 
Future work can also study the impact of \pga{} on physical interpretability of neural networks, since the features extracted at the hidden layers of the network correspond to physically meaningful concepts.